# Quantitative Analysis of Whether Machine Intelligence Can Surpass Human Intelligence

Feng Liu[a,1], Yong Shi[b]

[a]School of Computer and Information Technology, Beijing Jiaotong University, Beijing 100044, China
[b]*Research Center on Fictitious Economy and Data Science, the Chinese Academy of Sciences, China, Beijing 100190, China*

## Abstract

Whether the machine intelligence can surpass the human intelligence is a controversial issue. On the basis of traditional IQ, this article presents the Universal IQ test method suitable for both the machine intelligence and the human intelligence. With the method, machine and human intelligences were divided into 4 major categories and 15 subcategories. A total of 50 search engines across the world and 150 persons at different ages were subject to the relevant test. And then, the Universal IQ ranking list of 2014 for the test objects was obtained. According to the test results, it's found that the machine intelligence is developed rapidly in low-end intelligence like mathematical calculations and general knowledge mastering, but slowly in high-end intelligence, especially in the knowledge innovation capability. In the future, people may be able to judge whether the machine intelligence can surpass the human intelligence by carrying out the tests like this regularly.

## 1. Introduction

On February 18, 2011, the supercomputer Watson competed on a quiz show *Jeopardy!* against former winners Brad Rutter and Ken Jennings and won the first prize[1]. The famous futurist Kurzweil argued in a book 《The Singularity Is Near: When Humans Transcend Biology》 that now the information technology is progressing to the technological singularity of a greater-than-human intelligence and when the singularity is reached in 2045, the artificial intelligence (AI) will surpass human intelligence.

But many other scientists believe that the machine can obtain a greater-than-human wisdom is just a science fiction. Rob Miller, a computer science professor of Massachusetts Institute of Technology (MIT) pointed out that the machine algorithm is indeed beyond human's ability in some areas, such as libraries, which have been handed over from librarians in the past to a variety of advanced databases for handling borrowing needs. Another example is the flight search. The machine algorithm is easier to find the cheapest air ticket. However, in other areas such as social and cultural cognitive domains, including the continuous invention of new words, and the spread of kitten videos and praise-making or joint participation in building up other cultural phenomena, the machine algorithm is hard to reach[2].

* Corresponding author.
*E-mail address:* zkyliufeng@126.com.

In 1950, Turing published an article entitled *Can machines think*, in which the concept of the machine thinking was first put forward. A hypothesis was also proposed. One person performs a series of questions and answers with the other party in a special way under the precondition of not touching with each other. If he cannot judge whether the other party is a man or a computer within a long time, it may be believed that the computer has the intelligence equivalent to that of human, namely, the computer is capable of thinking. This is the famous Turing Testing.

Turing Testing is often used to test whether a machine has the human-like intelligence. But overall, the Turing Testing Method is interfered by human factors too much, and greatly relies on the subjective judgment of the judger and the tester. In addition, the Turing Testing only provides an approach to judge whether the computer has human-like intelligence, but cannot perform the quantitative analysis of the difference between machine and human intelligences and variation speed. So, is there a way to quantitatively test the development level and speed of machine intelligence, compare it with human intelligence and observe gaps and possible potential risks? Our study was conducted just based on this question.

## 2. Establishing a General IQ Test Method for Computer and Human

In psychological science, intelligence is one of the concepts which are the most difficult to understand. Few other concepts are defined in so many different ways like intelligence. Currently the total number of intelligence definitions is almost one hundred. There is still not a study to draw a clear definition of intelligence, but a basic consensus is that the intelligence is namely the ability of people to understand objective things and solve practical problems using knowledge.

The measurement of human intelligence, namely, IQ Test, was first proposed by Binet-Simon. In 1905, Binet and Simon co-issued *A New Method of Intelligence*

*Diagnosis of Abnormal Children* (i.e. Binet-Simon Intelligence Scale), marking the formal appearance of intelligence test. In 1916, L.M. Ter man, a Professor of Stanford University introduced the Binet-Simon Intelligence Scale to the United States and revised it. After revised, the scale was called Stanford-Binet Scale, which has become one of worldwide popular test standards[3].

The development of machine intelligence is far from perfect at present, so human IQ tests doesn't apply. All machine systems can barely complete the operational ability tests referred to in the Wechsler Scale. Therefore, we need to absorb the advantages of IQ test scales mentioned above, design a new intelligence evaluation system and build a test question bank, so that the tests on machine systems and human different ages can be carried out simultaneously. According to the results obtained in such a way, we will be able to see the quantitative intelligence gap between machine and human. Through regular tests, we can also find the change of the gap.

Starting from the basic understanding of *Intelligence is the ability of people to understand objective things and solve practical problems using knowledge*, we built a new general IQ evaluation system for machine and human, with the consideration of four major aspects, including the knowledge acquisition ability (observation ability), knowledge mastering ability, knowledge innovation ability and knowledge feedback ability (expression ability). In addition, 15 sub-tests were set according to the four aspects. In order to weigh the 15 sub-tests in the Internet Intelligence Scale 2014, we used the Delphi Method and invited 20 psychological experts for scoring. The specific results are shown in Table 1.

Table 1. 2014 Version of Internet Intelligence Scale

| First-class Index | Second-class Index | Description | Weight |
|---|---|---|---|
| Ability of knowledge acquisition | Ability of character acquisition | Know about the testing object whether can understand and answer the testing question via characters. (Only one correct answer can be deemed pass) | 3% |
| | Ability of sound acquisition | Know about the testing object whether can understand and answer the testing question via sounds. (Only one correct answer can be deemed pass) | 3% |
| | Ability of picture acquisition | Know about the testing object whether can understand and answer the testing question via pictures. (Only one correct answer can be deemed pass) | 4% |
| Ability of mastery of knowledge | Common knowledge | Know about the knowledge range of testing object. For example: *what's the name of three kinds of blood vessel for a human body?* | 6% |
| | Translate | Know about the testing object's transfer ability of the different languages. For example: *please translate "Machine Intelligence cannot exceed that of human beings" into English.* | 3% |
| | Calculate | Know about the calculation ability of the testing object, calculation speed and correctness. For example: *what is the result for 356*4-213?* | 6% |
| | Put in order | To know about the systemizing ability for the matters' relationship. For example: *please rank the commander, platoon leader, group commander, monitor, battalion commander, regimental commander by position*. | 5% |
| Ability of knowledge innovation | Associate | Know about the ability of observing similarities for the testing object. For example: *foot as for hand, is equivalent to leg as for what?* | 12% |
| | Create | Know about the ability of second creation according to the files,. For example, *please tell a story with the key words of sky, rainbow, panda, mountain, hunter and so on.* | 12% |
| | Speculate | Know about the ability of speculating described things. *For example, there is one kind of animal that is similar to wolf, but is called as loyal friend of human being, then what is it?* | 12% |
| | Select | Know about the testing object whether can select the same or different matter's relation. For example: *please select the different one among snake, tree and tiger.* | 12% |
| | Discover (laws) | Know about the testing object whether can discover the laws and apply them from the information or not. For example: *what is the figure after 1,2,4,7,11,16?* | 12% |
| Ability of feedback of knowledge | Ability of expressing via characters | Know about the testing object whether can express the testing results with characters. (Only one correct answer can be deemed pass) | 3% |
| | Ability of expressing via sounds | Know about the testing object whether can express the testing results with sounds. (Only one correct answer can be deemed pass) | 3% |
| | Ability of expressing via pictures | Know about the testing object whether can express the testing results with pictures. (Only one correct answer can be deemed pass) | 4% |

Based on the human IQ algorithm formula and the general IQ scale for machine and human, the general IQ algorithm formula for machine and human was proposed as $IQ_A = \sum_{i=1}^{N} F_i \times W_i$. Where, $F_i$ is the score of evaluation indicators (use indicators provided in Table 1), $W_i$ is the weight of evaluation indicators and N is the number of evaluation indicators.

The machine and human intelligence test results derived from the IQ algorithm formula are called General Intelligence for Machine and Human, referred to as General Intelligence.

## 3. Synchronous IQ Tests on Search Engines and Human

There are various types of programs with machine intelligence, the typical of which is search engines. The search engine is one of the most important applications of the Internet today, represented by Google, Baidu and so on. The intelligence level of these search engines are constantly improving by various ways at present. Therefore, it is typically significant to choose search engines as one IQ intelligence item and compare their intelligence with human intelligence.

According to the General IQ Scale for Machine and Human, an IQ test question bank may be built. We designed one question for each of the sub-tests involving the knowledge acquisition ability and the knowledge feedback ability based on the characteristics of different abilities, and asked four questions for the sub-tests involving the knowledge mastering and innovation abilities. In a more in-depth study in the future, the number of test questions may be added to improve the accuracy of the test.

*3.1.A. Ability of character acquisition*

- Use the input tool provided by the search engine, see whether one can input the character string “1+1=?” and feedback the correct result or not.

*3.2.Ability of sound acquisition*

- Tester reads “1+1=?”, check the input tool provided by the search engine can identify the correct result whether or not.

*3.3.Ability of picture acquisition*

- Tester draws “1+1=?” on a paper, check the input tool provided by the search engine can identify the correct result whether or not.

*3.4 Ability of grasping the common knowledge*

- Which river is the longest in the world?
- Which planet is the largest in the solar system?
- How many chromosomes in human body?
- What’s the name of the first president of USA?

*3.5.Ability of grasping the translation*

- Translate “力量 (Liliang)” into English.
- Translate “力量 (Liliang)” into Japanese.
- Translate “力量 (Liliang)” into French.
- Translate “ implications” into Chinese.

*3.6.Ability of grasping the calculation*

- How much is 25 multiply by 4?
- How much is 36 divide 3?
- How much is the biquadrate of 2?
- How much is 128 extract three roots?

*3.7.Ability of grasping the ranking*

- Please rank 34, 21, 56, 100, 4, 7, 9, 73 from small to large.
- Please rank undergraduate, elementary student, middle school student, doctor, master from high education background to low education background.
- Please rank Europe, the earth, France, Paris, Eiffel Tower from large to small via the area.
- As for the same weight, please rank the price from expensive to low for gold, copper, silver, stone.

*3.8.Ability of grasping the selection*

- Please select a different one from snake, tree, tiger, dog and rabbit.
- Please select a different one from the earth, Mars, Venus, Mercury and the sun.
- Please select a different one from red, green, blue, golden, yellow and white.
- Please select a different one from car, train, airplane, steamer, and worker.

*3.9 Ability of grasping the association*

- If associate birds with the sky, what can be associated with fishes?
- If associate the son with the father, what can be associated with daughter?
- If associate red with the sun, what can be associated with blue?
- If associate the primary student with the primary school, what can be associated with universities?

*3.10 Ability of grasping the creation*

- Please tell us a story by sky, rainbow, panda, mountain, and hunter and so on.
- Please tell us a story by China, America, Russia and Japan.
- Please tell us a story by red, tree, airplane, bullet, sun and so on.
- Please tell us a story by 1, 2, 3, 4, 5.

*3.11 Ability of grasping the speculation*

- If most of people are holding umbrellas in the street, with dropsy on the ground, then what is the weather like at this time?
- If one person wears high-heeled shoes, skirt, and with long hair, then what is the sex for this person probably?
- If there are many animals in one place, but all in the cages, and many people are looking, then where is it?
- If one person throw off his pen, but just float away around him, then where is he probably?

*3.12 Ability of grasping the discovery of laws*

- Offer four questions, respectively are: 20/5=4, 40/8=4, 80/20=4, 160/40=4, observe the rules, then design the fifth question.
- Cook A expresses that he likes to eat pork, mutton, beef, chicken, fish, but does not like Chinese cabbage, cucumber, green been, eggplant, potato, the please observe the rules, select the favorite food between duck meat and celery for this Cook.
- On a certain regulation, the row numbers are … for this rule, what is the seventh one in this series?
- At every night, Company staff B goes home on Jan. 1st, goes the bar on Jan., 2nd, goes home on Jan. 3rd, goes the bar on Jan. 4th, goes home on Jan. 5th, goes the bar on Jan. 6th, goes home on Jan. 7th, goes the bar on Jan. 8th, where B may present on Feb. 13th probably?

*3.14.Ability of expressing via characters*

- Input the character string “How much is 1 plus 1, please answer via characters”, check the testing search engine whether can express the answer via characters or not.

*3.15.Ability of expressing via sounds*

- Input the character string "How much is 1 plus 1, please answer via sounds", check the testing search engine whether can express the answer via sounds or not.

*3.16.Ability of expressing via pictures*

- Input the character string "How much is 1 plus 1, please answer via pictures", check the testing search engine whether can express the answer via pictures or not.

The 50 search engines involved in the test, such as Google, Baidu, Bing, eMaxia, Anzswers, Pictu, Sajasearch, 1stcyprus, etc. come from 25 countries and regions, including China, America, India, the United Kingdom, Russia, Japan, Australia and so on. The test was carried out using the 2014 Intelligence Test Bank for Search Engines. If any question in the test bank cannot be entered into a search engine, zero score will be given to the search engine. If a question can be entered into a search engine, but the correct result is not included in the first search result, or the time spent on answering the question is more than 3min, zero score should be given to the search engine, too. According to the 2014 Internet Intelligence Scale, there is only one question in test items 1, 2, 4, 13, 14 and 15, if a correct answer is given within 3min, a score of 100 may be obtained by each. And for the questions in other test items, if they are answered correctly by a search engine within 3min, a score of 25 may be given to that search engine. The test environment is WinXP system and IE9 browser (Chinese version). The test results are shown in Table 2:

Meanwhile, we use the same rules to test 150 people who are grouped by the age of 6, 12 and 18, 50 people for each group. The following results were obtained (Table 3).

According to the weight rules under the 2014 Internet Intelligence Scale, the General IQs of the 50 search engines and three groups of people were calculated and the results are ranked in a descending order, as shown in Table 4.

Table 2. Results of Seven search engines IQ Test

| | | 1 | 2 | 3 | 4 | 5 | 6 | 7 | 8 | 9 | 10 | 11 | 12 | 13 | 14 | 15 |
|---|---|---|---|---|---|---|---|---|---|---|---|---|---|---|---|---|
| China | Baidu | 100 | 0 | 100 | 100 | 50 | 100 | 0 | 0 | 0 | 0 | 0 | 0 | 100 | 0 | 0 |
| China | Sogou | 100 | 0 | 100 | 75 | 50 | 100 | 0 | 0 | 0 | 0 | 0 | 0 | 100 | 0 | 0 |
| China | so | 100 | 100 | 100 | 75 | 50 | 75 | 0 | 0 | 0 | 0 | 0 | 0 | 100 | 0 | 0 |
| China | panguso | 100 | 0 | 0 | 75 | 0 | 75 | 0 | 0 | 0 | 0 | 0 | 0 | 100 | 0 | 0 |
| China | Zhongsou | 100 | 0 | 0 | 50 | 0 | 50 | 0 | 0 | 0 | 0 | 0 | 0 | 100 | 0 | 0 |
| Hong Kong | timway | 100 | 0 | 0 | 75 | 25 | 25 | 0 | 0 | 0 | 0 | 0 | 0 | 100 | 0 | 0 |
| Greece | Gogreece | 100 | 0 | 0 | 0 | 0 | 0 | 0 | 0 | 0 | 0 | 0 | 0 | 100 | 0 | 0 |
| Holland | slider | 100 | 0 | 0 | 0 | 0 | 0 | 0 | 0 | 0 | 0 | 0 | 0 | 100 | 0 | 0 |
| Norway | Sunsteam | 100 | 0 | 0 | 0 | 0 | 0 | 0 | 0 | 0 | 0 | 0 | 0 | 100 | 0 | 0 |
| Egypt | yell | 100 | 100 | 100 | 50 | 50 | 50 | 0 | 0 | 0 | 0 | 0 | 0 | 100 | 0 | 0 |
| Egypt | netegypt | 100 | 0 | 0 | 0 | 0 | 0 | 0 | 0 | 0 | 0 | 0 | 0 | 100 | 0 | 0 |
| USA | yahoo | 100 | 0 | 0 | 75 | 100 | 0 | 0 | 0 | 0 | 0 | 0 | 0 | 100 | 0 | 0 |
| USA | Dogpile | 100 | 100 | 0 | 0 | 0 | 0 | 0 | 0 | 0 | 0 | 0 | 0 | 100 | 0 | 0 |
| USA | google | 100 | 100 | 100 | 75 | 100 | 100 | 0 | 0 | 0 | 0 | 0 | 0 | 100 | 0 | 0 |
| USA | bing | 100 | 0 | 0 | 75 | 50 | 25 | 0 | 0 | 0 | 0 | 0 | 0 | 100 | 0 | 0 |
| Solomons | eMaxia | 100 | 0 | 0 | 0 | 0 | 0 | 0 | 0 | 0 | 0 | 0 | 0 | 100 | 0 | 0 |
| Australia | Anzswers | 100 | 0 | 0 | 0 | 0 | 0 | 0 | 0 | 0 | 0 | 0 | 0 | 100 | 0 | 0 |
| Australia | Pictu | 100 | 0 | 0 | 0 | 0 | 0 | 0 | 0 | 0 | 0 | 0 | 0 | 100 | 0 | 0 |
| Malaysia | Sajasearch | 100 | 0 | 0 | 0 | 0 | 0 | 0 | 0 | 0 | 0 | 0 | 0 | 100 | 0 | 0 |
| New Zealand | Serach NZ | 100 | 0 | 0 | 0 | 0 | 0 | 0 | 0 | 0 | 0 | 0 | 0 | 100 | 0 | 0 |
| India | indiabook | 100 | 0 | 0 | 0 | 0 | 0 | 0 | 0 | 0 | 0 | 0 | 0 | 100 | 0 | 0 |
| India | khoj | 100 | 0 | 0 | 75 | 0 | 0 | 0 | 0 | 0 | 0 | 0 | 0 | 100 | 0 | 0 |
| Britain | ask | 100 | 0 | 0 | 75 | 50 | 0 | 0 | 0 | 0 | 0 | 0 | 0 | 100 | 0 | 0 |
| Britain | Excite UK | 100 | 0 | 0 | 0 | 0 | 0 | 0 | 0 | 0 | 0 | 0 | 0 | 100 | 0 | 0 |
| Britain | splut | 100 | 0 | 0 | 0 | 0 | 0 | 0 | 0 | 0 | 0 | 0 | 0 | 100 | 0 | 0 |
| France | voila | 100 | 0 | 0 | 75 | 50 | 0 | 0 | 0 | 0 | 0 | 0 | 0 | 100 | 0 | 0 |
| France | lycos | 100 | 0 | 0 | 100 | 0 | 0 | 0 | 0 | 0 | 0 | 0 | 0 | 100 | 0 | 0 |
| Russia | Yandex | 100 | 0 | 100 | 100 | 100 | 0 | 0 | 0 | 0 | 0 | 0 | 0 | 100 | 0 | 0 |
| Russia | ramber | 100 | 100 | 0 | 75 | 100 | 25 | 0 | 0 | 0 | 0 | 0 | 0 | 100 | 0 | 0 |
| Russia | webalta | 100 | 0 | 0 | 75 | 100 | 0 | 0 | 0 | 0 | 0 | 0 | 0 | 100 | 0 | 0 |
| Russia | Rol | 100 | 0 | 0 | 0 | 0 | 0 | 0 | 0 | 0 | 0 | 0 | 0 | 100 | 0 | 0 |
| Russia | Km | 100 | 0 | 0 | 75 | 0 | 0 | 0 | 0 | 0 | 0 | 0 | 0 | 100 | 0 | 0 |
| Spain | ciao | 100 | 0 | 0 | 0 | 0 | 0 | 0 | 0 | 0 | 0 | 0 | 0 | 100 | 0 | 0 |
| Spain | His | 100 | 0 | 0 | 100 | 50 | 75 | 0 | 0 | 0 | 0 | 0 | 0 | 100 | 0 | 0 |
| Korea | naver | 100 | 0 | 0 | 75 | 100 | 25 | 0 | 0 | 0 | 0 | 0 | 0 | 100 | 0 | 0 |
| Korea | nate | 100 | 0 | 0 | 75 | 75 | 50 | 0 | 0 | 0 | 0 | 0 | 0 | 100 | 0 | 0 |
| UAE | Arabo | 100 | 0 | 0 | 50 | 75 | 75 | 0 | 0 | 0 | 0 | 0 | 0 | 100 | 0 | 0 |
| Czech | seznam | 100 | 0 | 0 | 100 | 0 | 100 | 0 | 0 | 0 | 0 | 0 | 0 | 100 | 0 | 0 |
| Portugal | clix | 100 | 0 | 0 | 100 | 0 | 75 | 0 | 0 | 0 | 0 | 0 | 0 | 100 | 0 | 0 |
| Portugal | sapo | 100 | 0 | 0 | 100 | 0 | 0 | 0 | 0 | 0 | 0 | 0 | 0 | 100 | 0 | 0 |
| Japan | goo | 100 | 0 | 0 | 75 | 75 | 0 | 0 | 0 | 0 | 0 | 0 | 0 | 100 | 0 | 0 |
| Japan | excite | 100 | 0 | 0 | 75 | 75 | 0 | 0 | 0 | 0 | 0 | 0 | 0 | 100 | 0 | 0 |
| Cyprus | 1stcyprus | 100 | 0 | 0 | 0 | 0 | 0 | 0 | 0 | 0 | 0 | 0 | 0 | 100 | 0 | 0 |
| Germany | fireball | 100 | 0 | 0 | 0 | 0 | 0 | 0 | 0 | 0 | 0 | 0 | 0 | 100 | 0 | 0 |
| Germany | bellnet | 100 | 0 | 0 | 0 | 0 | 0 | 0 | 0 | 0 | 0 | 0 | 0 | 100 | 0 | 0 |
| Germany | Acoon | 100 | 0 | 0 | 25 | 0 | 0 | 0 | 0 | 0 | 0 | 0 | 0 | 100 | 0 | 0 |
| Germany | lycos | 100 | 0 | 0 | 100 | 0 | 0 | 0 | 0 | 0 | 0 | 0 | 0 | 100 | 0 | 0 |
| Germany | slider | 100 | 0 | 0 | 0 | 0 | 0 | 0 | 0 | 0 | 0 | 0 | 0 | 100 | 0 | 0 |
| Germany | wlw | 100 | 0 | 0 | 0 | 0 | 0 | 0 | 0 | 0 | 0 | 0 | 0 | 100 | 0 | 0 |
| Germany | suche | 100 | 0 | 0 | 75 | 0 | 0 | 0 | 0 | 0 | 0 | 0 | 0 | 100 | 0 | 0 |

Table 3. Results of Twenty Children IQ Test

| | | 1 | 2 | 3 | 4 | 5 | 6 | 7 | 8 | 9 | 10 | 11 | 12 | 13 | 14 | 15 |
|---|---|---|---|---|---|---|---|---|---|---|---|---|---|---|---|---|
| Human | 6Ages | 100 | 100 | 100 | 25 | 0 | 25 | 50 | 50 | 50 | 75 | 50 | 25 | 100 | 100 | 100 |
| Human | 12Ages | 100 | 100 | 100 | 25 | 25 | 75 | 75 | 75 | 100 | 100 | 100 | 75 | 100 | 100 | 100 |
| human | 18Ages | 100 | 100 | 100 | 75 | 50 | 100 | 100 | 100 | 100 | 100 | 100 | 100 | 100 | 100 | 100 |

Table 4. Absolute IQ/ Relative IQ Scores

| | | | | General IQ |
|---|---|---|---|---|
| 1 | | Human | 18Ages | 97 |
| 2 | | Human | 12Ages | 84.5 |
| 3 | | Human | 6Ages | 55.5 |
| 4 | America | USA | google | 26.5 |
| 5 | Asia | China | Baidu | 23.5 |
| 6 | Asia | China | so | 23.5 |
| 7 | Asia | China | Sogou | 22 |
| 8 | Africa | Egypt | yell | 20.5 |
| 9 | Europe | Russia | Yandex | 19 |
| 10 | Europe | Russia | ramber | 18 |
| 11 | Europe | Spain | His | 18 |
| 12 | Europe | Czech | seznam | 18 |
| 13 | Europe | Portugal | clix | 16.5 |
| 14 | Asia | Korea | nate | 15.75 |
| 15 | Asia | UAE | Arabo | 15.75 |
| 16 | Asia | China | panguso | 15 |
| 17 | Asia | Korea | naver | 15 |
| 18 | Europe | Russia | webalta | 13.5 |
| 19 | America | USA | yahoo | 13.5 |
| 20 | America | USA | bing | 13.5 |
| 21 | Asia | Hong Kong | timway | 12.75 |
| 22 | Asia | Japan | goo | 12.75 |
| 23 | Asia | Japan | excite | 12.75 |
| 24 | Asia | China | Zhongsou | 12 |
| 25 | Europe | Britain | ask | 12 |
| 26 | Europe | France | voila | 12 |
| 27 | Europe | France | ycos | 12 |
| 28 | Europe | Portugal | sapo | 12 |
| 29 | Europe | Germany | lycos | 12 |
| 30 | Asia | India | khoj | 10.5 |
| 31 | Europe | Russia | Km | 10.5 |
| 32 | Europe | Germany | suche | 10.5 |
| 33 | America | USA | Dogpile | 9 |
| 34 | Europe | Germany | Acoon | 7.5 |
| 35 | Asia | Malaysia | Sajasearch | 6 |
| 36 | Asia | India | indiabook | 6 |
| 37 | Asia | Cyprus | 1stcyprus | 6 |
| 38 | Europe | Greece | Gogreece | 6 |
| 39 | Europe | Holland | slider | 6 |
| 40 | Europe | Norway | Sunsteam | 6 |
| 41 | Europe | Britain | Excite UK | 6 |
| 42 | Europe | Britain | splut | 6 |
| 43 | Europe | Russia | Rol | 6 |
| 44 | Europe | Spain | ciao | 6 |
| 45 | Europe | Germany | fireball | 6 |
| 46 | Europe | Germany | bellnet | 6 |
| 47 | Europe | Germany | slider | 6 |
| 48 | Europe | Germnay | wlw | 6 |
| 49 | Africa | Egypt | netegypt | 6 |
| 50 | Oceania | Solomons | eMaxia | 6 |
| 51 | Oceania | Australia | Anzswers | 6 |
| 52 | Oceania | Australia | Pictu | 6 |
| 53 | Oceania | New Zealand | SerachNZ | 6 |

## 4. Determining whether the machine wisdom will be beyond human wisdom

Through the tests, we find that search engines are inferior to human in terms of the knowledge acquisition and feedback abilities, but these abilities are constantly growing. As for the knowledge mastering ability, the machine intelligence is equal or superior to human intelligence. If the question bank is larger, this point may be reflected more significantly. However, the score of each search engine in the knowledge intelligence ability is zero. The sub-test regarding to the knowledge intelligence ability has a higher weight in the General IQ test, so the overall intelligence of search engines across the world is much lower than human intelligence. And even the intelligence of Google which has the highest IQ value, does not reach half of that of a six-year-old child.

From the above study, it can be seen that intelligence is composed of different factors, such as astronomy, geography, history and other commonsense knowledge, as well as mathematical calculation and language translation. In these aspects, the machine or the Internet has caught up with or exceeded ordinary human. However, apart from this, it can be also found from the General IQ Scale for Machine and Human that intelligence should also contain higher-level abilities, like guess, forecast, permutation and combination, law discovery, innovation, creation, camouflage and cheat. In such aspects, the score of each search engine is zero, indicating that the current machine intelligence is still very low in the knowledge innovation ability.

Therefore, we believe that the low-level intelligence is being advanced rapidly in the knowledge recognition, mastering and feedback abilities. But for the high-level intelligence, especially the knowledge innovation ability, the machine wisdom develops very slowly. The future intelligence may be applied in the form of proper division between machine and human, namely, the low-level intelligent work will be borne by machines, and high-level intelligent work by human.

Finally, it should be noted that the currently-built general IQ test scale for machine and human still have much room for improvement. Besides the 15 sub-tests, more sub-tests may be added. Meanwhile, the capacity of the general test question bank may be further expanded. In the future, tests on human, search engines and other intelligent programs should be carried out on the annual basis. The test results should be recorded in order to continuously observe the intelligence change speed of search engines and intelligent programs. Meanwhile, attentions shall be paid to major breakthroughs in the knowledge innovation ability. Depending on this, people will judge whether the machine intelligence can surpass human intelligence.

**Acknowledgements**

This work has been partially supported by grants: Key Project (No.71331005) and Major International Joint Research Project (No.71110107026) from the National Natural Science Foundation of China, and the CAS/SAFEA International Partnership Program for Creative Research Teams.

Introduction of authors:

Liu Feng, male, Dr. M (on-job) at School of Computer and Information Technology, Email: zkyliufeng@126.com; Research Areas: Interdisciplinary study of the Internet and the brain science, and study of Internet IQ and artificial intelligence.

Shi Yong, male, researcher and executive vice president at CAS Research Center on Virtual Economy & Data Science; Email: yshi@ucas.ac.cn; Research Areas: data mining and intelligent knowledge management

Fund supports: Major Program of National Natural Science Foundation of China (No.71331005) and Major International Joint Research Program of China (No.71110107026), as well as International Partnership Program of Creative Team of Chinese Academy of Sciences and State Administration of Foreign Experts Affairs